\documentclass{article} 
\usepackage{iclr2015,times}
\usepackage{hyperref}
\usepackage{url}
\usepackage{natbib}
\usepackage{amsmath}
\usepackage{amssymb}
\usepackage{graphicx}
\usepackage{subfig}
\usepackage{wrapfig}

\title{An Analysis of Unsupervised Pre-training in Light of Recent Advances}

\author{
Tom Le Paine*, Pooya Khorrami*, Wei Han, Thomas S. Huang \\ \thanks{ - Authors contributed equally to this work.}
Beckman Institute for Advanced Science and Technology\\
University of Illinois at Urbana-Champaign\\
Urbana, IL 61801\\
\texttt{{paine1,pkhorra2,weihan3,t-huang1}@illinois.edu} \\
}

\iclrfinalcopy 

\begin{document}

\maketitle

\begin{abstract}
Convolutional neural networks perform well on object recognition because of a number of recent advances: rectified linear units (ReLUs), data augmentation, dropout, and large labelled datasets. Unsupervised data has been proposed as another way to improve performance. Unfortunately, unsupervised pre-training is not used by state-of-the-art methods leading to the following question: Is unsupervised pre-training still useful given recent advances? If so, when? We answer this in three parts: we 1) develop an unsupervised method that incorporates ReLUs and recent unsupervised regularization techniques, 2) analyze the benefits of unsupervised pre-training compared to data augmentation and dropout on CIFAR-10 while varying the ratio of unsupervised to supervised samples, 3) verify our findings on STL-10. 
We discover unsupervised pre-training, as expected, helps when the ratio of unsupervised to supervised samples is high, and surprisingly, hurts when the ratio is low. We also use unsupervised pre-training with additional color augmentation to achieve near state-of-the-art performance on STL-10.

\end{abstract}

\section{Introduction}
\label{sec:intro}
We analyze the benefits of unsupervised pre-training in the context of recent deep learning innovations including: rectified linear units, data augmentation, and dropout. Recent work shows that convolutional neural networks (CNNs) can achieve state-of-the-art performance for object classification (\citet{krizhevsky2012imagenet}) and object detection (\citet{girshick2013rich}), when there is enough training data. However, in many cases there is a dearth of labeled data. In these cases regularization is necessary for good results. The most common types of regularization are data augmentations (\citet{krizhevsky2012imagenet, dosovitskiy2014discriminative}) and dropout (\citet{hinton2012improving}). Another form of regularization, unsupervised pre-training (\citet{hinton2006fast, bengio2007greedy, erhan2010does}), has recently fallen out of favor. 

While there has been significant work in unsupervised learning, most of these works came before rectified linear units, which significantly help training deep supervised neural networks, and before simpler regularization schemes for unsupervised learning, such as zero-bias with linear encoding for auto-encoders (\citet{memisevic2014zero}).

We train an unsupervised method that takes advantage of these improvements we call Zero-bias Convolutional Auto-encoders (CAEs). Previous work showed that pre-trained tanh CAEs achieved an increase in performance over randomly initialized tanh CNNs. We conduct this experiment with our zero-bias CAE and observe a larger boost in performance. 

We analyze the effectiveness of our technique when combined with the popular regularization techniques used during supervised training on CIFAR-10 while varying the ratio of unsupervised to supervised samples. We do this comparing against randomly initialized CNNs without any additional regularization. We find that, when ratio is large, unsupervised pre-training provides useful regularization, increasing test set performance. When the ratio is small, we find that unsupervised pre-training hurts performance. 

We verify our finding that unsupervised pre-training can boost performance when the ratio of unsupervised to supervised samples is high by running our algorithm on the STL-10 dataset, which has a ratio of 100:1. As expected, we observe an improvement (3.87\%). When combined with additional color augmentation, we achieve near state-of-the-art results. Our unsupervised regularization still yields an improvement of (1.69\%).

We will begin by reviewing related work on fully-connected and convolutional auto-encoders. In Section \ref{sec:method}, we will present our method and how it is trained both during unsupervised pre-training and supervised fine-tuning. We present our results on the CIFAR-10 and STL-10 datasets in Section \ref{sec:experiments}, and in Section \ref{sec:conclusions} we conclude the paper.



\section{Related Work}
\label{sec:related_work}
Many methods have used unsupervised learning to learn parameters, which are subsequently used to initialize a neural network to be trained on supervised data. These are called unsupervised pre-training, and supervised fine-tuning respectively. We will highlight some of the unsupervised learning methods related to our work.

\subsection{Auto-encoders}

One of the most widely-used models for unsupervised learning, an auto-encoder is a model that learns a function that minimizes the squared error between the input $x \in \mathbb{R}^n$ and its reconstruction $r(x)$:
\begin{gather}
L = \| x - r(x) \|_{2}^{2} \\
r(x) = W_{d}^{T}f(W_{e}x+b)+c
\label{eq:ae_recon}
\end{gather}
In the above equation, $W_{e}$ represents the weight matrix that transforms the input, $x$ into some hidden representation, $b$ is vector of biases for each hidden unit and $f(\cdot)$ is some nonlinear function. Commonly chosen examples for $f(\cdot)$ include the sigmoid and hyperbolic tangent functions. Meanwhile, $W_{d}$ is the weight matrix that maps back from the hidden representation to the input space and $c$ is a vector of biases for each input (visible) unit. These parameters are commonly learned by minimizing the loss function over the training data via stochastic gradient descent.

When no other constraints are imposed on the loss function, the auto-encoder weights tend to learn the identity function. To combat this, some form of regularization must imposed upon the model so that the model can uncover the underlying structure in the data. Some forms of regularization include adding noise to the input units (\citet{vincent2010stacked}) and requiring the hidden unit activations be sparse (\citet{coates2011analysis}) or have small derivatives (\citet{rifai2011contractive}). These models are known as de-noising, sparse, and contractive auto-encoders respectively. A more recent work by \citet{memisevic2014zero} showed that training an auto-encoder with rectified linear units (ReLU) caused the activations to form tight clusters due to having negative bias values. They showed that using thresholded linear (TLin) or thresholded rectifier (TRec) activations with no bias can allow one to train an auto-encoder without the need for additional regularization.

\subsection{Convolutional Auto-encoders}

While the aforementioned fully-connected techniques have shown impressive results, they do not directly address the structure of images. Convolutional neural networks (CNNs) (\citet{lecun1998gradient, lee2009convolutional}) present a way to reduce the number of connections by having each hidden unit only be responsible for a small local neighborhood of visible units. Such schemes allow for dense feature extraction followed by pooling layers which when stacked could allow the network to learn over larger and larger receptive fields. Convolutional auto-encoders (CAEs) combined aspects from both auto-encoders and convolutional neural nets making it possible to extract highly localized patch-based information in an unsupervised fashion. There have been several works in this area including \citet{jarrett2009best} and \citet{zeiler2010deconvolutional}. Both rely on sparse coding to force their unsupervised learning to learn non-trival solutions. \citet{zeiler2011adaptive} extended this work by introducing pooling/unpooling and visualizing how individual feature maps at different layers influenced specific portions of the reconstruction. These sparse coding approaches had limitations because they used an iterative procedure for inference. A later work by \citet{masci2011stacked} trained deep feed forward convolutional auto-encoders, using only max-pooling and saturating tanh non-linearities as a form of regularization, while still showing a modest improvement over randomly initialized CNNs. While tanh was a natural choice at the time, \citet{krizhevsky2012imagenet} showed that ReLUs are more suitable for learning given their non-saturating behavior.

\section{Our Approach}
\label{sec:method}
Our method's training framework can be broken up into two phases: (i) unsupervised pre-training and (ii) supervised fine-tuning. We describe those in more detail below.

\subsection{Unsupervised Pre-training}
Our method incorporates aspects of previous unsupervised learning methods in order to learn salient features, yet be efficient to train. Our model is similar to the deconvolutional network in \citet{zeiler2011adaptive} where the cost we minimize at each layer is the mean square error on the original image. However, unlike the network in \citet{zeiler2011adaptive}, our method does not use any form of sparse coding. Our model also is similar to that of \citet{masci2011stacked}, however we improve upon it by introducing regularization in the convolutional layers through the use of zero-biases and ReLUs as discussed in \citet{memisevic2014zero}. 

We now describe the model architecture in detail. Like the previous work described above, our model involves several encoding modules followed by several decoding modules. A single encoding module $E_{l}(\cdot)$ consists of a convolution layer $F_{l}$, a nonlinearity $f(\cdot)$, followed by a pooling layer $P_{s_l}$ with switches $s_l$.

\begin{equation}
E_{l}(x) = P_{s_l}f(F_{l}x)
\label{eq:encoder}
\end{equation}

Each encoding module has an associated decoding module $D_{l}$, which unpools using $E_{l}$ pooling switches $s_l$ and deconvolves with $E_{l}$'s filters, (i.e. $F_{l}^T$).

\begin{equation}
D_{l}(x) = F_{l}^TU_{s_l}x
\label{eq:decoder}
\end{equation}

A two layer network can be written as:

\begin{equation}
r(x) = D_{1}(D_{2}(E_{2}(E_{1}(x))))
\label{eq:reconstruction}
\end{equation}

We train each encoder/decoder pair in a greedy fashion (i.e. first a 1 layer CAE, then a 2 layer CAE, etc.) while keeping the parameters of previous layers fixed. Like \citet{zeiler2011adaptive}, we compute the cost by taking the mean squared error between the original image and the network's reconstruction of the input. Thus, the costs for a one layer network ($C_1(x)$) and two layer network ($C_2(x)$) would be expressed in the following manner:
\begin{gather}
C_1(x) = \| x - D_{1}(E_{1}(x)) \|_{2}^{2} \\
C_2(x) = \| x - D_{1}(D_{2}(E_{2}(E_{1}(x)))) \|_{2}^{2}
\label{eq:layer_cost}
\end{gather}

We regularize our learned representation by fixing the biases of our convolutional and deconvolutional layers at zero and using ReLUs as our activation function during encoding. We use linear activations for our decoders. Unlike the work by \citet{memisevic2014zero} which analyzes fully-connected auto-encoders, our work is the first, to our knowledge, that trains zero-bias CAEs for unsupervised learning.

\subsubsection{Unsupervised Weight Initialization}
Weight initialization is often a key component of successful neural network training. For ReLU's it is important to ensure the input to the ReLU is greater than 0. This can be achieved by setting the bias appropriately. This cannot be done for zero-bias auto-encoders. Instead we use two methods for initializing the weights to achieve this 1)  in the first layer, we initialize each of the filters to be a randomly drawn patch from the dataset, 2) on the later layers, we sample weights from a Gaussian distribution and find the nearest orthogonal matrix by taking the singular value decomposition (SVD) of the weight matrix and setting all of the singular values to one. For CNNs we must take into account the additive effect of overlapping patches thus we weight each filter by a 2D hamming window to prevent intensity build-up.

\subsection{Supervised Fine-tuning}
After the weights of the CAE have been trained, we remove all of the decoder modules and leave just the encoding modules. We add an additional fully-connected layer and a softmax layer to the pre-trained encoding modules. The weights of these layers are drawn from a Gaussian distribution with zero mean and standard deviation of $ k/\sqrt{N_{FAN\_IN}}$, where $k$ is drawn uniformly from $\left[0.2, 1.2\right]$. 

\subsection{Training}
For both unsupervised and supervised training we use stochastic gradient descent with a constant  momentum of 0.9, and a weight decay parameter of 1e-5. We select the highest learning rate that doesn't explode for the duration of training. For these experiments we do not anneal the learning rate. The only pre-processing we do to each patch is centering (i.e. mean subtraction) and scaling to unit variance.

\section{Experiments and Analysis}
\label{sec:experiments}
\subsection{Datasets}
We run experiments on two natural image datasets, CIFAR-10 (\cite{krizhevsky2009learning}) and STL-10 (\cite{coates2011analysis}). CIFAR-10 is a common benchmark for object recognition. Many unsupervised and supervised neural network approaches have been tested on it. It consists of 32x32 pixel color images drawn from 10 object categories. It has 50,000 training images, and 10,000 testing images. STL-10 is also an object recognition benchmark, but was designed to test unsupervised learning algorithms, so it has a relatively small labeled training set of 500 images per class, and an additional unsupervised set which contains 100,000 unlabeled images. The test set contains 800 labeled images per class. All examples are 96x96 pixel color images.

\subsection{CIFAR-10}

On CIFAR-10, we train a network with structure similar to \citet{masci2011stacked}, so that we can directly show the benefits of our modifications. The network consists of three convolutional layers with 96, 144, and 192 filters respectively. The filters in the first two layers are of size 5x5 while the filters in the third layer are of size 3x3. We also add 2x2 max pooling layers after the first two convolutional layers. There is also a full-connected layer with 300 hidden units followed by a softmax layer with 10 output units. All of our nets were trained using our own open source neural network library \footnote{\url{https://github.com/ifp-uiuc/anna}}.

As stated in the methods section, we first train our unsupervised model on 100\% of the training images, do supervised fine-tuning, and report overall accuracy on the test set. We 1) present qualitative results of unsupervised learning, 2) show our zero-bias convolutional auto-encoder performs well compared to previous convolutional auto-encoder work by \citet{masci2011stacked} developed before the popularization of rectified linear units, and zero-bias auto-encoders, 3) we show our analysis of various regularization techniques, and vary the ratio of unsupervised to supervised data, 4) for completeness we report our best results when training on the full CIFAR-10 dataset, however this is not the main point of this work.

\subsubsection{Qualitative Results}
\begin{wrapfigure}{R}{0.5\textwidth}
\centering
\includegraphics[width=0.48\textwidth]{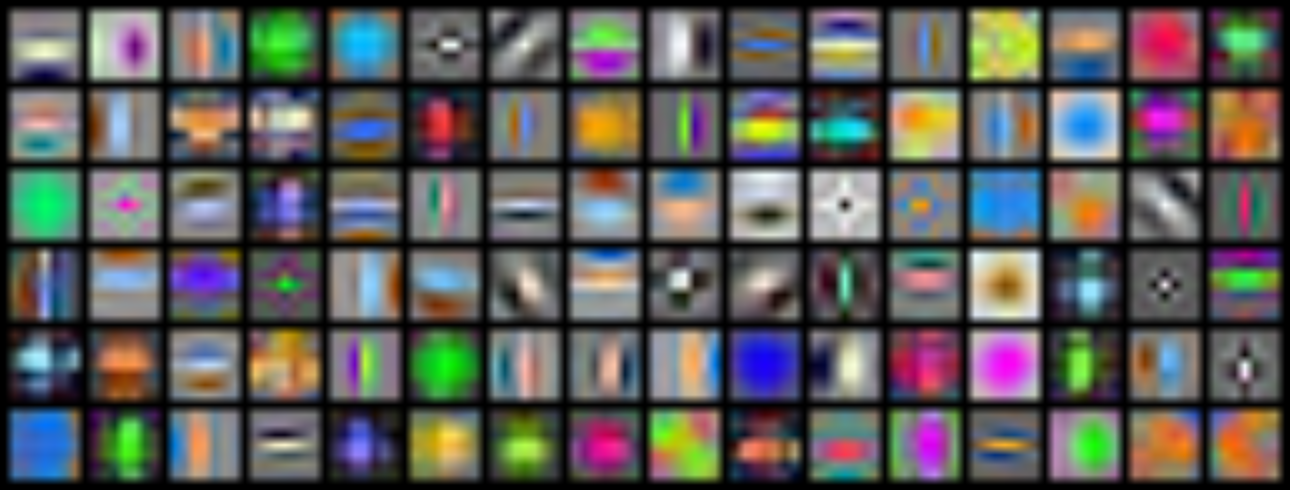}
\caption{First layer filters learned by our zero-bias convolutional auto-encoder. Each filter has dimension 7x7x3. (Best viewed in color.) For direct comparison with tanh CAE please see \citet{masci2011stacked} Figure 2c.}
\label{fig:layer1_filters}
\vspace{-0.5cm}
\end{wrapfigure}

One way in which we ensure the quality of our learned representation is by inspecting the first layer filters. We visualize the filters learned by our model in Figure \ref{fig:layer1_filters}. So that we can directly compare with the filters presented in \citet{masci2011stacked}, we trained an additional zero-bias convolutional auto-encoder with filters of size 7x7x3 (instead of 5x5x3) in the first layer. From Figure \ref{fig:layer1_filters}, we can see that, indeed, our model is able to capture interpretable patterns such as Gabor-like oriented edges (both color and intensity) and center-surrounds.

\subsubsection{Unsupervised Pre-training for Tanh CAEs and Zero-bias CAEs}

For our quantitative experiments, we first compare the performance of the tanh CAE proposed by \citet{masci2011stacked} with our zero-bias CAE. In their paper, \citet{masci2011stacked} trained a tanh CNN from a random initialization and compared it with one pre-trained using a tanh CAE. They also added 5\% translations as a form of data augmentation. We re-conduct this experiment using a zero-bias CNN trained from a random initialization, and compare it to one pre-trained using our zero-bias CAE.

In Table \ref{tab:masci_comp} we compare the improvements of our model with that of \citet{masci2011stacked}'s, on various subsets of CIFAR-10. As expected, the zero-bias CNN (a ReLU CNN without bias parameters) performs significantly better than the tanh CNN (2.53\%, 8.53\%, 5.23\%). More interestingly, notice that on each subset, compared to \citet{masci2011stacked} our pre-trained model shows similar or better performance over the randomly initialized CNN.  When the ratio of unsupervised to supervised data is high, we experience an 8.44\% increase in accuracy as opposed to \citet{masci2011stacked}'s 3.22\% increase.

\begin{table}[b]
\begin{center}
    \caption{Comparison between Tanh CAE (\citet{masci2011stacked}) and our model on various subsets of CIFAR-10.}
    \label{tab:masci_comp}
    \begin{tabular}{ |l | c | c | c | c |}
    \hline
    \multicolumn{1}{|p{5cm}|}{\raggedright Unsupervised to supervised ratio \\ (Samples per Class)}
& \multicolumn{1}{|p{1cm}|}{\centering 50:1 \\ (100)}
& \multicolumn{1}{|p{1cm}|}{\centering 10:1 \\ (500)}
& \multicolumn{1}{|p{1cm}|}{\centering 5:1  \\ (1000)}
& \multicolumn{1}{|p{1cm}|}{\centering 1:1  \\ (5000)} \\ \hline
     Tanh CNN - \citet{masci2011stacked} & 44.48 \% & --- & 64.77 \% & 77.50 \% \\ 
     Tanh CAE - \citet{masci2011stacked} & 47.70 \% & --- & 65.65 \% & 78.20 \% \\ \hline
     Zero-bias CNN  & 47.01 \% & 64.76 \%& 73.30 \% & 82.73 \% \\
     \bf Zero-bias CAE  & \bf 55.45 \% & \bf 68.42 \% & \bf 74.06 \% & \bf 83.64 \% \\ \hline
    \end{tabular}
\end{center}
\end{table}

\subsubsection{Analysis of regularization methods}
\begin{figure}[t]
\centering
\subfloat[ 50:1 unsupervised to supervised sample ratio, (100 samples per class), baseline CNN: 44.3\%][50:1 unsupervised to supervised sample ratio \\ (100 samples per class), baseline CNN: 44.3\%]{
\includegraphics[trim=0.5cm 0cm 0cm 0cm, clip=true, scale=0.475]{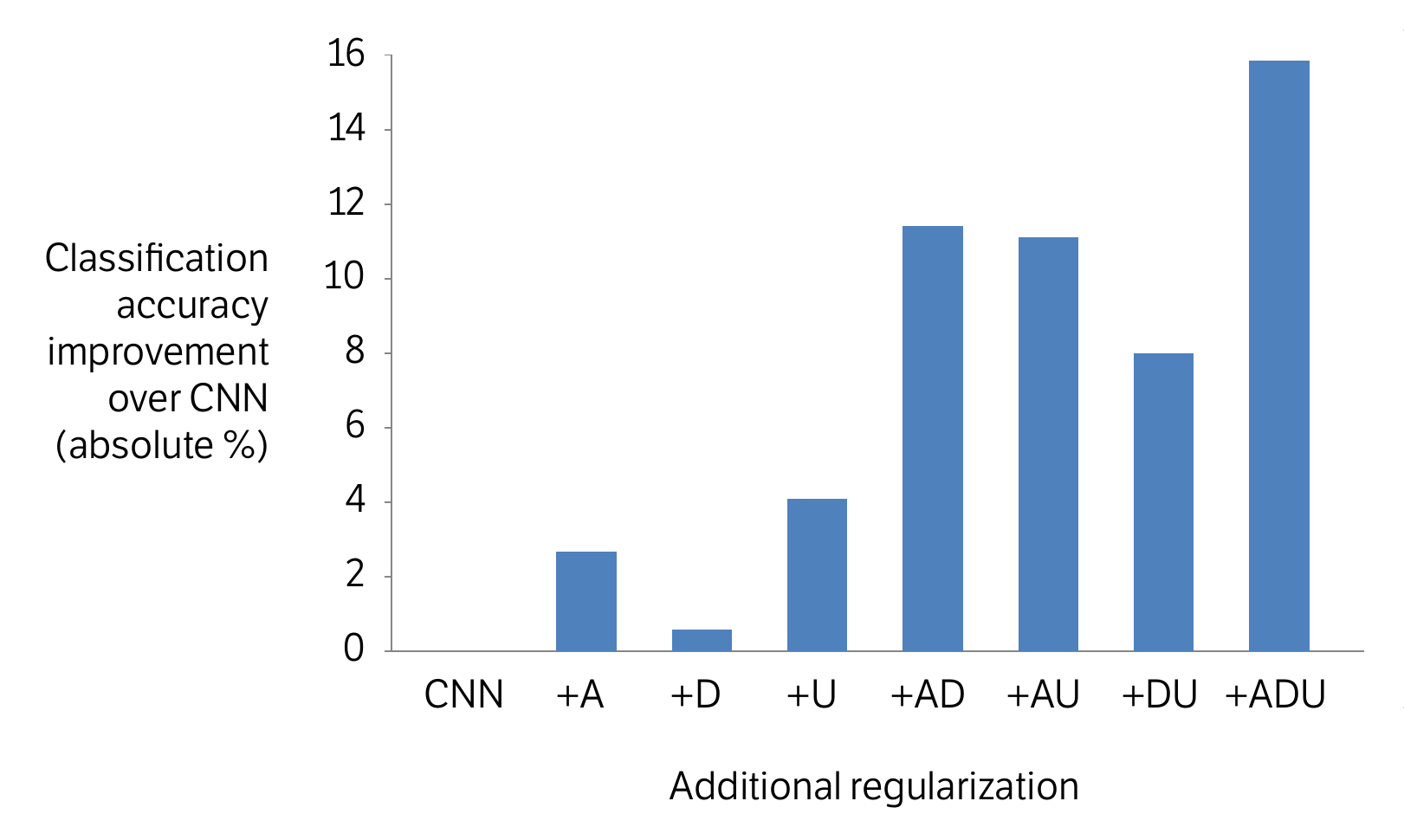}
\label{fig:100_bar}}
\subfloat[10:1 unsupervised to supervised sample ratio, (500 samples per class), baseline CNN: 62.0\%][10:1 unsupervised to supervised sample ratio\\ (500 samples per class), baseline CNN: 62.0\%]{
\includegraphics[trim=3.2cm 0cm 0cm 0cm, clip=true, scale=0.475]{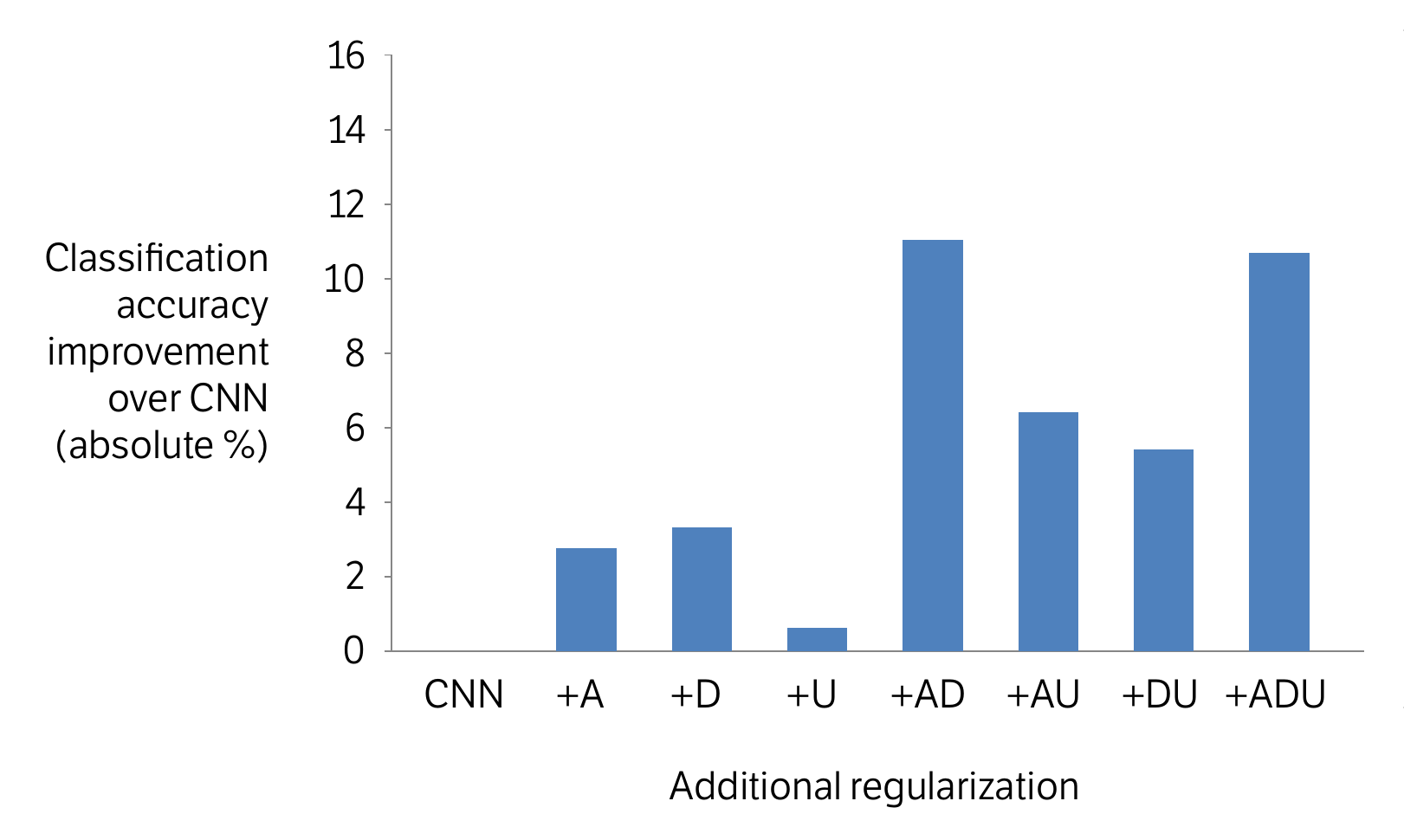}
\label{fig:500_bar}}
\\
\subfloat[5:1 unsupervised to supervised sample ratio, (1000 samples per class), baseline CNN: 67.8\%][5:1 unsupervised to supervised sample ratio\\ (1000 samples per class), baseline CNN: 67.8\%]{
\includegraphics[trim=0cm 0cm 0cm 0cm, scale=0.475]{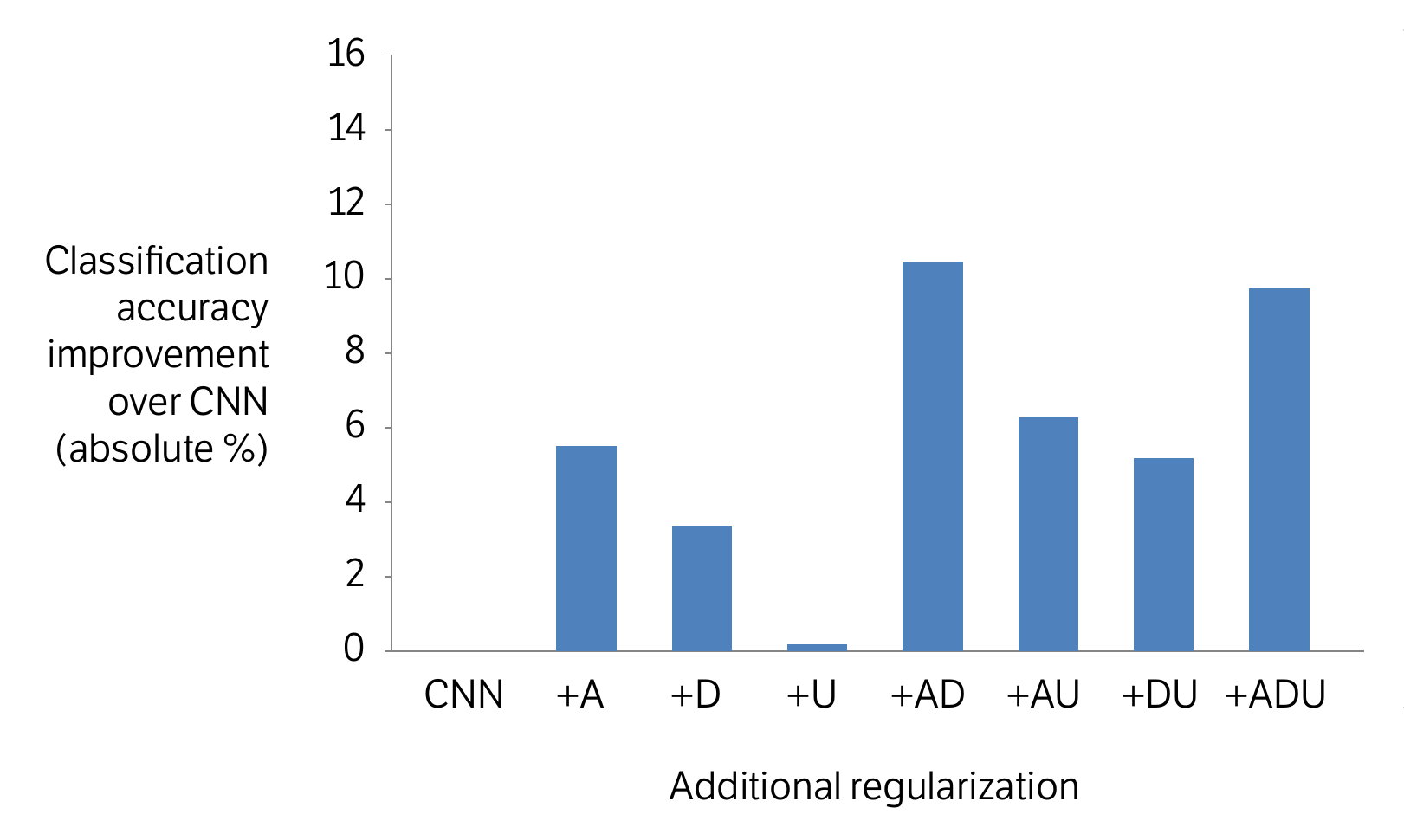}
 \label{fig:1000_bar}}
\subfloat[1:1 unsupervised to supervised sample ratio, (5000 samples per class), baseline CNN: 80.2\%][1:1 unsupervised to supervised sample ratio\\ (5000 samples per class), baseline CNN: 80.2\%]{
\includegraphics[trim=3.2cm 0cm 0cm 0cm, clip=true, scale=0.475]{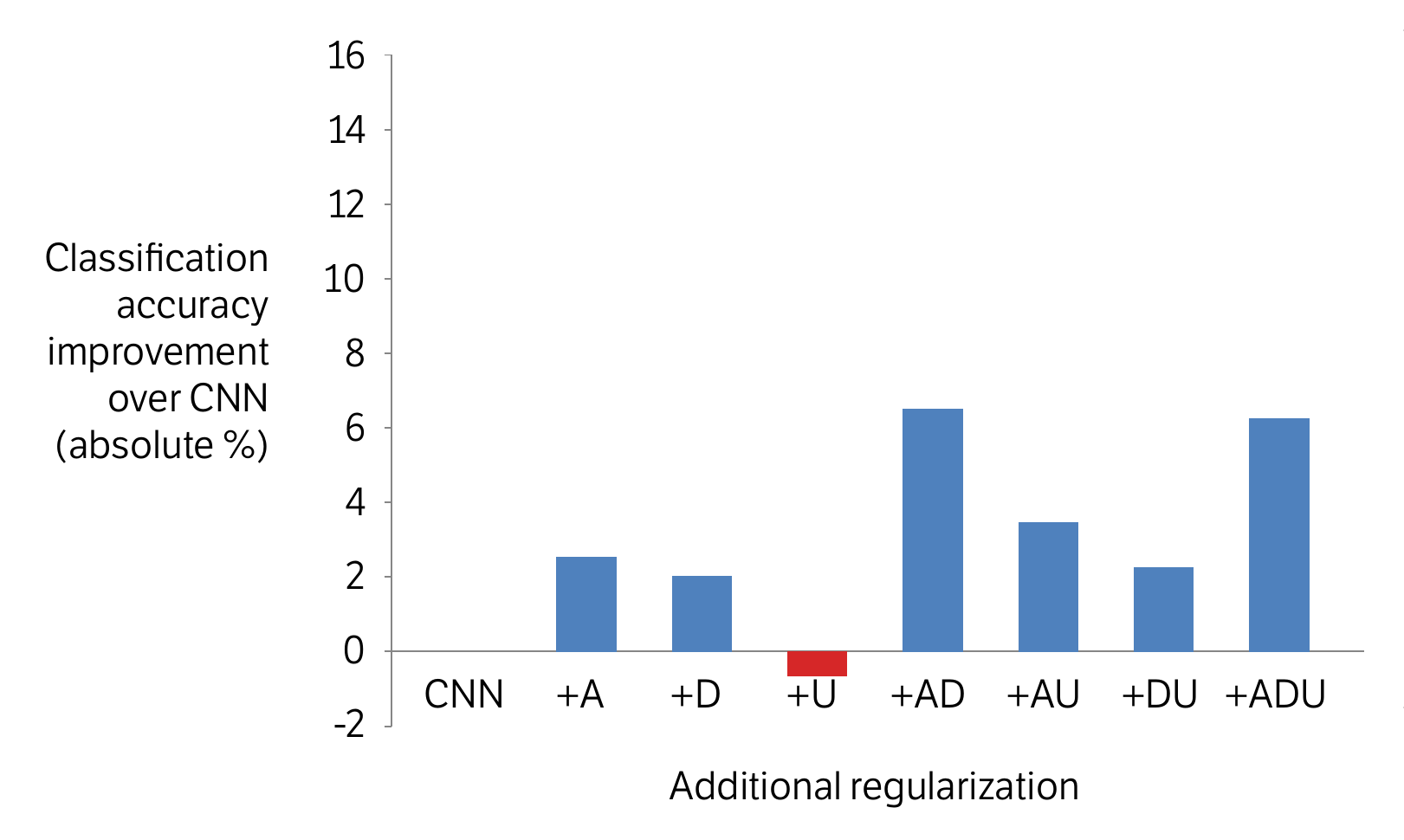}
\label{fig:5000_bar}}
\caption{Analysis of the effects of different types of regularization (A: data augmentation, D: dropout, U: unsupervised learning), individually and jointly, on different subsets of CIFAR-10.}
\label{fig:improvement_bar}
\vspace{-0.5cm}
\end{figure}

Next, we analyze how different supervised regularization techniques affect our model's performance. Specifically, we consider the effects of dropout, data augmentation (via translations and horizontal flips), unsupervised pre-training (with our zero-bias CAE) and their combinations. We compare each regularization technique to a zero-bias CNN trained from random initialization without any regularization (labeled CNN in Figure \ref{fig:improvement_bar}). Figure \ref{fig:improvement_bar} shows the classification accuracy improvement over CNN for each type of regularization both individually and together.

We perform this analysis for subsets of CIFAR-10 with different unsupervised to supervised sample ratios ranging from 50:1 to 1:1, by fixing the unsupervised data size, and varying the number of supervised examples. It is important to note that as this ratio approaches 1:1, the experimental setup favors data augmentation and dropout because the number of virtual supervised samples is larger than number of unsupervised samples.

In Figure \ref{fig:100_bar}, where the ratio of unsupervised to supervised samples is 50:1, there are three notable effects: (i) unsupervised pre-training alone yields a larger improvement (4.09\%) than data augmentation (2.67\%) or dropout (0.59\%), (ii) when unsupervised pre-training is combined with either data augmentation or dropout, the improvement is greater than the sum of the individual contributions, (iii) we experience the largest gains (15.86\%) when we combine all three forms of regularization.

We see that effect (ii) is also observed in the case where the ratio of unsupervised to supervised samples is 10:1 (Figure \ref{fig:500_bar}), and to a lesser extent when the ratio is 5:1 (Figure \ref{fig:1000_bar}). Unfortunately, effects (i) and (iii) are not observed when the ratio of unsupervised to supervised samples decreases. We will elaborate on effect (i) below.

\begin{wrapfigure}{R}{0.55\textwidth}
\vspace{-0.6cm}
\centering
\includegraphics[trim = 0.75cm 0cm 0cm 0cm , clip=true, width=0.53\textwidth]{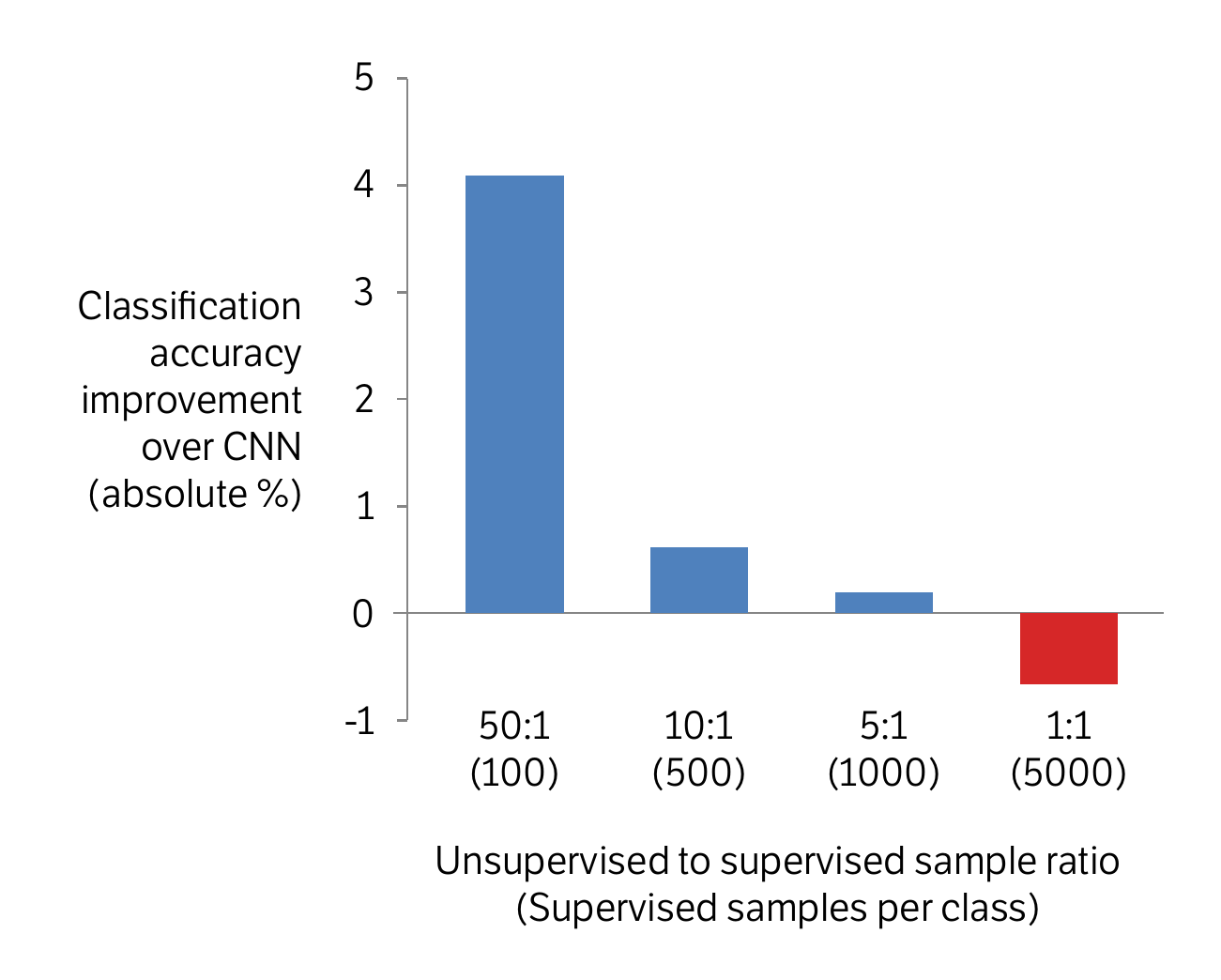}
\caption{The benefits of unsupervised learning vs. unsupervised to supervised sample ratio. When the ratio is 50:1, we see a 4.09\% increase in performance. But the benefit shrinks as the ratio decreases. When the ratio is 1:1, there is a penalty for using unsupervised pre-training.}
\label{fig:decay_plot}
\vspace{-1.2cm}
\end{wrapfigure}

In Figure \ref{fig:decay_plot}, we observe that the improvement in performance from unsupervised learning decreases rapidly as the ratio of unsupervised to supervised samples decreases. Surprisingly, when the ratio is 1:1, we see that unsupervised learning actually hurts performance (-0.67\%).

\subsubsection{Comparison with existing methods}

We also compare the performance of our algorithm on the full CIFAR-10 dataset with other techniques in Table \ref{tab:method_comp_cifar}, though we show above our method performs worse when the ratio of unsupervised to supervised samples is 1:1. We outperform all methods that use unsupervised pre-training (\citet{masci2011stacked}, \citet{coates2011analysis}, \citet{dosovitskiy2014discriminative}, \citet{lin2014stable}), however we are not competitive with supervised state-of-the-art. We include some representative supervised methods in Table \ref{tab:method_comp_cifar}. 

\begin{table}[b]
\begin{center}
    \caption{Quantitative comparison with other methods on CIFAR-10 (A: Data Augmentation, D: Dropout, U: Unsupervised Learning).}
    \label{tab:method_comp_cifar}
    \begin{tabular}{ |l | c |}
    \hline
    Algorithm & Accuracy \\ \hline    
    Convolutional Auto-encoders - \citet{masci2011stacked} & 79.20 \% \\ 
    Single layer K-means - \citet{coates2011analysis} & 79.60 \% \\ 
    Convolutional K-means Networks - \citet{coates2011selecting} & 82.00 \% \\ 
    Exemplar CNN - \citet{dosovitskiy2014discriminative} & 82.00 \% \\ 
    Convolutional Kernel Networks - \citet{Mairal2014convolutional} & 82.18 \% \\ 
    NOMP - \citet{lin2014stable} & 82.90 \% \\ 
    Max-Out Networks - \citet{goodfellow2013maxout} & 90.65 \% \\
    Network In Network - \citet{lin2013network} & 91.20 \% \\ 
    \textbf{Deeply-Supervised Nets - \citet{lee2014deeply}} & \textbf{91.78 \%} \\
    \hline
    \hline
    Zero-bias CNN +ADU &  86.44 \% \\ \hline
    Zero-bias CNN +AD &  86.70 \% \\ 
    \hline
    \end{tabular}
\end{center}
\end{table}

\subsection{STL-10}

Next, we assess the effects of unsupervised pre-training on STL-10. From the CIFAR-10 experiments, it is clear unsupervised pre-training can be beneficial if the unsupervised dataset is much larger than the supervised dataset. STL-10 was designed with this in mind, and has a ratio of unsupervised to supervised data of 100:1. So we experimentally show this benefit. 

We design our network to have structure similar to \citet{dosovitskiy2014discriminative}, to ease comparison. The network used consists 3 convolutional layers with 64, 128, and 256 filters in each layer, a fully-connected layer with 512 units, and a softmax layer with 10 output units. We also apply max-pooling layers of size 2x2 after the first two convolutional layers and quadrant pooling after the third convolutional layer.

We train the zero-bias CAE on 100,000 unlabeled images. We then fine-tune the network on each of the 10 provided splits of training set, each consisting of 1000 samples (100 samples per class), and evaluate all of them on the test set. The accuracies are subsequently averaged to obtain the final recognition accuracy. Similar to our CIFAR-10 experiments, we also train a zero-bias CNN with the same structure as our zero-bias CAE on each of the splits to further highlight the benefits of unsupervised learning.

Table \ref{tab:method_comp_stl} presents our results on the STL-10 dataset and compares them with other methods. As expected, unsupervised pre-training gives a 3.87\% increase over the randomly initialized CNN.

\subsubsection{Additional data augmentation: color and contrast}
The current best result on STL-10 (\citet{dosovitskiy2014discriminative}) makes extensive use of additional data augmentation including: scaling, rotation, color and two forms of contrast. They do not perform these augmentations during supervised training, but during a discriminative unsupervised feature learning period. We test the regularizing effects of these additional augmentations when applied directly to supervised training, and test how these regularization effects hold up when combined with with unsupervised pre-training. To do this, we use some of these additional data-augmentations during our supervised training: \textbf{color augmentation} and \textbf{contrast augmentation}.

\textbf{Color augmentation:}
The images are represented in HSV color space (h, s, v). Here we generate a single random number for each image and add it to the hue value for each pixel like so:
\begin{eqnarray}
a &\sim& Uniform(-0.1, 0.1) \\
h &=& h + a
\end{eqnarray}

\textbf{Contrast augmentation:} 
Here we generate six random numbers for each image, with the following distributions:
\begin{eqnarray}
a, d &\sim& Uniform(0.7, 1.4) \\
b, e &\sim& Uniform(0.25, 4) \\
c, f &\sim& Uniform(-0.1, 0.1)
\end{eqnarray}

And use them to modify the saturation and value for every pixel in the image, like so:
\begin{eqnarray}
s &=& a s^b + c \\
v &=& d s^e + f
\end{eqnarray}

We find that a) additional data-augmentation is incredibly helpful, increasing accuracy by 6.5\%, b) unsupervised pre-training still maintains an advantage (1.69\%).

\begin{table}
\begin{center}
    \caption{Quantitative comparison with other methods on STL-10 (A: Data Augmentation, D: Dropout, C: Color Augmentation, U: Unsupervised Learning).}
    \label{tab:method_comp_stl}
    \begin{tabular}{ |l | c |}
    \hline
    Algorithm & Accuracy \\ \hline    
    Convolutional K-means Networks - \citet{coates2011selecting} & 60.1 \% $\pm$ 1.0 \% \\ 
    Convolutional Kernel Networks - \citet{Mairal2014convolutional} & 62.32 \% \\ 
    Hierarchical Matching Pursuit (HMP) - \citet{bo2013unsupervised} & 64.5 \% $\pm$ 1.0 \% \\ 
    NOMP - \citet{lin2014stable} & 67.9 \% $\pm$ 0.6 \% \\
    Multi-task Bayesian Optimization - \citet{swersky2013multi} & 70.1 \% $\pm$ 0.6 \% \\ 
    \textbf{Exemplar CNN - \citet{dosovitskiy2014discriminative}} & \textbf{72.8 \% $\pm$ 0.4 \% }\\ \hline
    \hline
     Zero-bias CNN +AD &  62.01 \% $\pm$ 1.9 \% \\ \hline
     Zero-bias CNN +ADU &  65.88 \% $\pm$ 0.9 \% \\ \hline
     Zero-bias CNN +ADC & 68.51 \% $\pm$ 0.8 \% \\ \hline
     Zero-bias CNN +ADCU & 70.20 \% $\pm$ 0.7 \% \\ 
    \hline
    \end{tabular}
\end{center}
\end{table}

\section{Conclusions}
\label{sec:conclusions}
We present a new type of convolutional auto-encoder that has zero-bias and ReLU activations and achieves superior performance to previous methods.
We conduct thorough experiments on CIFAR-10 to analyze the effects of unsupervised pre-training as a form of regularization when used in isolation and in combination with supervised forms of regularization such as data augmentation and dropout. We observe that, indeed, unsupervised pre-training can provide a large gain in performance when the ratio of unsupervised to supervised samples is large. 
Finally, we verify our findings by applying our model to STL-10, a dataset with far more unlabeled samples than labeled samples (100:1). We find that with additional regularization, via color augmentation, our method is able to achieve nearly state-of-the-art results.

\subsubsection*{Code}
All experiments were run using our own open source library Anna, which can be found at: \url{https://github.com/ifp-uiuc/anna}

Code to reproduce the experiments can be found at: \url{https://github.com/ifp-uiuc/an-analysis-of-unsupervised-pre-training-iclr-2015}

\subsubsection*{Acknowledgments}
This material is based upon work supported by the National Science Foundation under Grant No. 392 NSF IIS13-18971. The two Tesla K40 GPUs used for this research were donated by the NVIDIA Corporation. We would like to acknowledge Theano (\citet{bergstra2010theano}) and Pylearn2 (\citet{goodfellow2013pylearn2}), on which our code is based. Also, we would like to thank Shiyu Chang for many helpful discussions and suggestions.

\bibliography{main}

\begin{thebibliography}{28}
\providecommand{\natexlab}[1]{#1}
\providecommand{\url}[1]{\texttt{#1}}
\expandafter\ifx\csname urlstyle\endcsname\relax
  \providecommand{\doi}[1]{doi: #1}\else
  \providecommand{\doi}{doi: \begingroup \urlstyle{rm}\Url}\fi

\bibitem[Bengio et~al.(2007)Bengio, Lamblin, Popovici, Larochelle,
  et~al.]{bengio2007greedy}
Yoshua Bengio, Pascal Lamblin, Dan Popovici, Hugo Larochelle, et~al.
\newblock Greedy layer-wise training of deep networks.
\newblock \emph{Advances in neural information processing systems},
  19:\penalty0 153, 2007.

\bibitem[Bergstra et~al.(2010)Bergstra, Breuleux, Bastien, Lamblin, Pascanu,
  Desjardins, Turian, Warde-Farley, and Bengio]{bergstra2010theano}
James Bergstra, Olivier Breuleux, Fr{\'{e}}d{\'{e}}ric Bastien, Pascal Lamblin,
  Razvan Pascanu, Guillaume Desjardins, Joseph Turian, David Warde-Farley, and
  Yoshua Bengio.
\newblock Theano: a {CPU} and {GPU} math expression compiler.
\newblock In \emph{Proceedings of the Python for Scientific Computing
  Conference ({SciPy})}, June 2010.
\newblock Oral Presentation.

\bibitem[Bo et~al.(2013)Bo, Ren, and Fox]{bo2013unsupervised}
Liefeng Bo, Xiaofeng Ren, and Dieter Fox.
\newblock Unsupervised feature learning for rgb-d based object recognition.
\newblock In \emph{Experimental Robotics}, pages 387--402. Springer, 2013.

\bibitem[Coates and Ng(2011)]{coates2011selecting}
Adam Coates and Andrew~Y Ng.
\newblock Selecting receptive fields in deep networks.
\newblock In \emph{Advances in Neural Information Processing Systems}, pages
  2528--2536, 2011.

\bibitem[Coates et~al.(2011)Coates, Ng, and Lee]{coates2011analysis}
Adam Coates, Andrew~Y Ng, and Honglak Lee.
\newblock An analysis of single-layer networks in unsupervised feature
  learning.
\newblock In \emph{International Conference on Artificial Intelligence and
  Statistics}, pages 215--223, 2011.

\bibitem[Dosovitskiy et~al.(2014)Dosovitskiy, Springenberg, Riedmiller, and
  Brox]{dosovitskiy2014discriminative}
Alexey Dosovitskiy, Jost~Tobias Springenberg, Martin Riedmiller, and Thomas
  Brox.
\newblock Discriminative unsupervised feature learning with convolutional
  neural networks.
\newblock In Z.~Ghahramani, M.~Welling, C.~Cortes, N.D. Lawrence, and K.Q.
  Weinberger, editors, \emph{Advances in Neural Information Processing Systems
  27}, pages 766--774. Curran Associates, Inc., 2014.

\bibitem[Erhan et~al.(2010)Erhan, Bengio, Courville, Manzagol, Vincent, and
  Bengio]{erhan2010does}
Dumitru Erhan, Yoshua Bengio, Aaron Courville, Pierre-Antoine Manzagol, Pascal
  Vincent, and Samy Bengio.
\newblock Why does unsupervised pre-training help deep learning?
\newblock \emph{The Journal of Machine Learning Research}, 11:\penalty0
  625--660, 2010.

\bibitem[Girshick et~al.(2013)Girshick, Donahue, Darrell, and
  Malik]{girshick2013rich}
Ross Girshick, Jeff Donahue, Trevor Darrell, and Jitendra Malik.
\newblock Rich feature hierarchies for accurate object detection and semantic
  segmentation.
\newblock \emph{arXiv preprint arXiv:1311.2524}, 2013.

\bibitem[Goodfellow et~al.(2013{\natexlab{a}})Goodfellow, Warde-Farley,
  Lamblin, Dumoulin, Mirza, Pascanu, Bergstra, Bastien, and
  Bengio]{goodfellow2013pylearn2}
Ian~J Goodfellow, David Warde-Farley, Pascal Lamblin, Vincent Dumoulin, Mehdi
  Mirza, Razvan Pascanu, James Bergstra, Fr{\'e}d{\'e}ric Bastien, and Yoshua
  Bengio.
\newblock Pylearn2: a machine learning research library.
\newblock \emph{arXiv preprint arXiv:1308.4214}, 2013{\natexlab{a}}.

\bibitem[Goodfellow et~al.(2013{\natexlab{b}})Goodfellow, Warde-Farley, Mirza,
  Courville, and Bengio]{goodfellow2013maxout}
Ian~J Goodfellow, David Warde-Farley, Mehdi Mirza, Aaron Courville, and Yoshua
  Bengio.
\newblock Maxout networks.
\newblock \emph{arXiv preprint arXiv:1302.4389}, 2013{\natexlab{b}}.

\bibitem[Hinton et~al.(2006)Hinton, Osindero, and Teh]{hinton2006fast}
Geoffrey Hinton, Simon Osindero, and Yee-Whye Teh.
\newblock A fast learning algorithm for deep belief nets.
\newblock \emph{Neural computation}, 18\penalty0 (7):\penalty0 1527--1554,
  2006.

\bibitem[Hinton et~al.(2012)Hinton, Srivastava, Krizhevsky, Sutskever, and
  Salakhutdinov]{hinton2012improving}
Geoffrey~E. Hinton, Nitish Srivastava, Alex Krizhevsky, Ilya Sutskever, and
  Ruslan Salakhutdinov.
\newblock Improving neural networks by preventing co-adaptation of feature
  detectors.
\newblock \emph{CoRR}, abs/1207.0580, 2012.

\bibitem[Jarrett et~al.(2009)Jarrett, Kavukcuoglu, Ranzato, and
  LeCun]{jarrett2009best}
Kevin Jarrett, Koray Kavukcuoglu, M~Ranzato, and Yann LeCun.
\newblock What is the best multi-stage architecture for object recognition?
\newblock In \emph{Computer Vision, 2009 IEEE 12th International Conference
  on}, pages 2146--2153. IEEE, 2009.

\bibitem[Krizhevsky and Hinton(2009)]{krizhevsky2009learning}
Alex Krizhevsky and Geoffrey Hinton.
\newblock Learning multiple layers of features from tiny images.
\newblock \emph{Computer Science Department, University of Toronto, Tech. Rep},
  2009.

\bibitem[Krizhevsky et~al.(2012)Krizhevsky, Sutskever, and
  Hinton]{krizhevsky2012imagenet}
Alex Krizhevsky, Ilya Sutskever, and Geoffrey~E Hinton.
\newblock Imagenet classification with deep convolutional neural networks.
\newblock In \emph{Advances in neural information processing systems}, pages
  1097--1105, 2012.

\bibitem[LeCun et~al.(1998)LeCun, Bottou, Bengio, and
  Haffner]{lecun1998gradient}
Yann LeCun, L{\'e}on Bottou, Yoshua Bengio, and Patrick Haffner.
\newblock Gradient-based learning applied to document recognition.
\newblock \emph{Proceedings of the IEEE}, 86\penalty0 (11):\penalty0
  2278--2324, 1998.

\bibitem[Lee et~al.(2014)Lee, Xie, Gallagher, Zhang, and Tu]{lee2014deeply}
Chen-Yu Lee, Saining Xie, Patrick Gallagher, Zhengyou Zhang, and Zhuowen Tu.
\newblock Deeply-supervised nets.
\newblock \emph{arXiv preprint arXiv:1409.5185}, 2014.

\bibitem[Lee et~al.(2009)Lee, Grosse, Ranganath, and Ng]{lee2009convolutional}
Honglak Lee, Roger Grosse, Rajesh Ranganath, and Andrew~Y Ng.
\newblock Convolutional deep belief networks for scalable unsupervised learning
  of hierarchical representations.
\newblock In \emph{Proceedings of the 26th Annual International Conference on
  Machine Learning}, pages 609--616. ACM, 2009.

\bibitem[Lin et~al.(2013)Lin, Chen, and Yan]{lin2013network}
Min Lin, Qiang Chen, and Shuicheng Yan.
\newblock Network in network.
\newblock \emph{arXiv preprint arXiv:1312.4400}, 2013.

\bibitem[Lin and Kung(2014)]{lin2014stable}
Tsung-Han Lin and H.~T. Kung.
\newblock Stable and efficient representation learning with nonnegativity
  constraints.
\newblock In Tony Jebara and Eric~P. Xing, editors, \emph{Proceedings of the
  31st International Conference on Machine Learning (ICML-14)}, pages
  1323--1331. JMLR Workshop and Conference Proceedings, 2014.

\bibitem[Mairal et~al.(2014)Mairal, Koniusz, Harchaoui, and
  Schmid]{Mairal2014convolutional}
Julien Mairal, Piotr Koniusz, Zaid Harchaoui, and Cordelia Schmid.
\newblock Convolutional kernel networks.
\newblock In Z.~Ghahramani, M.~Welling, C.~Cortes, N.D. Lawrence, and K.Q.
  Weinberger, editors, \emph{Advances in Neural Information Processing Systems
  27}, pages 2627--2635. Curran Associates, Inc., 2014.

\bibitem[Masci et~al.(2011)Masci, Meier, Cire{\c{s}}an, and
  Schmidhuber]{masci2011stacked}
Jonathan Masci, Ueli Meier, Dan Cire{\c{s}}an, and J{\"u}rgen Schmidhuber.
\newblock Stacked convolutional auto-encoders for hierarchical feature
  extraction.
\newblock In \emph{Artificial Neural Networks and Machine Learning--ICANN
  2011}, pages 52--59. Springer, 2011.

\bibitem[Memisevic et~al.(2014)Memisevic, Konda, and
  Krueger]{memisevic2014zero}
Roland Memisevic, Kishore Konda, and David Krueger.
\newblock Zero-bias autoencoders and the benefits of co-adapting features.
\newblock \emph{arXiv preprint arXiv:1402.3337}, 2014.

\bibitem[Rifai et~al.(2011)Rifai, Vincent, Muller, Glorot, and
  Bengio]{rifai2011contractive}
Salah Rifai, Pascal Vincent, Xavier Muller, Xavier Glorot, and Yoshua Bengio.
\newblock Contractive auto-encoders: Explicit invariance during feature
  extraction.
\newblock In \emph{Proceedings of the 28th International Conference on Machine
  Learning (ICML-11)}, pages 833--840, 2011.

\bibitem[Swersky et~al.(2013)Swersky, Snoek, and Adams]{swersky2013multi}
Kevin Swersky, Jasper Snoek, and Ryan~P Adams.
\newblock Multi-task bayesian optimization.
\newblock In \emph{Advances in Neural Information Processing Systems}, pages
  2004--2012, 2013.

\bibitem[Vincent et~al.(2010)Vincent, Larochelle, Lajoie, Bengio, and
  Manzagol]{vincent2010stacked}
Pascal Vincent, Hugo Larochelle, Isabelle Lajoie, Yoshua Bengio, and
  Pierre-Antoine Manzagol.
\newblock Stacked denoising autoencoders: Learning useful representations in a
  deep network with a local denoising criterion.
\newblock \emph{The Journal of Machine Learning Research}, 11:\penalty0
  3371--3408, 2010.

\bibitem[Zeiler et~al.(2010)Zeiler, Krishnan, Taylor, and
  Fergus]{zeiler2010deconvolutional}
Matthew~D Zeiler, Dilip Krishnan, Graham~W Taylor, and Robert Fergus.
\newblock Deconvolutional networks.
\newblock In \emph{Computer Vision and Pattern Recognition (CVPR), 2010 IEEE
  Conference on}, pages 2528--2535. IEEE, 2010.

\bibitem[Zeiler et~al.(2011)Zeiler, Taylor, and Fergus]{zeiler2011adaptive}
Matthew~D Zeiler, Graham~W Taylor, and Rob Fergus.
\newblock Adaptive deconvolutional networks for mid and high level feature
  learning.
\newblock In \emph{Computer Vision (ICCV), 2011 IEEE International Conference
  on}, pages 2018--2025. IEEE, 2011.

\end{thebibliography}
\bibliographystyle{plainnat}

\end{document}